# Restrained Generative Adversarial Network against Overfitting in Numeric Data Augmentation

Wei Wang [1,2], Yimeng Chai [1], Tao Cui [3], Chuang Wang [1], Baohua Zhang [1], Yue Li [1,2,*], Yi An [4,*]

[1] College of Computer Science, Nankai University, Tianjin 300350, China

[2] Key Laboratory for Medical Data Analysis and Statistical Research of Tianjin (KLMDASR), Tianjin 300350, China

[3] NCMIS, LSEC, Academy of Mathematics and Systems Science, Chinese Academy of Sciences, Beijing 100190, China

[4] Agro-environmental Protection Institute, Ministry of Agriculture, Tianjin 300071, China.

Corresponding author: Yue Li (e-mail: liyue80@nankai.edu.cn), Yi An (e-mail: simon8601@126.com).

## ABSTRACT

In recent studies, Generative Adversarial Network (GAN) is one of the popular schemes to augment the image dataset. However, in our study we find the generator $G$ in the GAN fails to generate numerical data in lower-dimensional spaces, and we address overfitting in the generation. By analyzing the Directed Graphical Model (DGM), we propose a theoretical restraint, independence on the loss function, to suppress the overfitting. Practically, as the Statically Restrained GAN (SRGAN) and Dynamically Restrained GAN (DRGAN), two frameworks are proposed to employ the theoretical restraint to the network structure. In the static structure, we predefined a pair of particular network topologies of $G$ and $D$ as the restraint, and quantify such restraint by the interpretable metric Similarity of the Restraint ($SR$). While for DRGAN we design an adjustable dropout module for the restraint function. In the widely carried out 20 group experiments, on four public numerical class imbalance datasets and five classifiers, the static and dynamic methods together produce the best augmentation results of 19 from 20; and both two methods simultaneously generate 14 of 20 groups of the top-2 best, proving the effectiveness and feasibility of the theoretical restraints.

## KEYWORDS

## 1 Introduction

Generative Adversarial Network (GAN) [1] is widely applied in the fields of computer vision, such as text to image [2] and image to image [3,4] translation and super-resolution [5]. Due to the dimensions of image space, the existing GANs setup a pair of network topologies on $G$ and $D$, within a large parameter-space. For example, the parameter set of the BigGAN is up to 1.6 billion [31]. Till now, it is still in the research to apply the GAN to the relatively low-dimensional numerical data, and a few research models are included in the GAN in 2014[12], WGAN in 2016[6] and GAN-DAE [13] in 2018. Our experiment found GAN-based augmentations for numerical data failed to show apparent advancement compared with the original data and SMOTE-augmented data. Take Table 1 for example. Table 1 shows that the traditional GANs cannot significantly improve AUC in RF classifiers in the four datasets and more experimental results on more classifiers can be found in section IV.

**Table 1** Failures of data augmentation by GANs (Evaluated by classifier RFC and the top-2 are bold)

|  | *Dateset[1]* | *Dataset[2]* | *Dataset[3]* | *Dataset[4]* |
|---|---|---|---|---|
| ***Techniques*** | | *AUC* | | |
| Baseline[1]-SMOTE [9] | **0.9201** | **0.7600** | **0.8052** | **0.8092** |
| Baseline[2]-Original data | **0.9207** | 0.7540 | 0.7927 | **0.7716** |
| GAN [1] | 0.9186 | 0.7359 | 0.7937 | 0.7634 |
| GAN-DAE [13] | 0.9161 | 0.7360 | 0.8042 | 0.7598 |
| WGAN [6] | 0.9147 | **0.7638** | 0.7755 | 0.7665 |

Note: All GAN-based methods generate worse argument data on the four datasets, evaluated by the classifier RFC, than the **Baseline[1]-SMOTE** and **Baseline[2]-Original data**. (**Dataset[1]**: Australian Credit Approval dataset, **Dataset[2]**: German Credit dataset, **Dataset[3]**: Pima Indians Diabetes dataset, **Dataset[4]**: SPECT Heart dataset)

In the machine learning theory, bad performance usually is blaming for either overfitting or underfitting. In the scene of generating structural data, we would instead address this poor generating to overfitting of generator $G$, with regards to the dimensional difference between dataset and the parameter set of $G$. Table 2 demonstrates the dimensions of the datasets and the number of parameters of GANs. Noted that understand a certain probabilistic distribution, i.e. Joint Gaussian Distribution, the DoF (degree of freedom) could be $O(n^2)$, which is dramatically smaller than the DoF of $G$ in GAN.

There are currently quite a few concerns about the overfitting of $G$ for both practical and theoretical reasons. Practically, in the majority of GAN's application, image generation, the dimension of the image space is form 0.43 million up to 1.6 billion and hence, concerning the DoF in image space, most GANs seem to be underfitting rather than overfitting when low-quality samples are generated. Theoretically, the DoF of the GAN is not quite clear till now, since the parameter of $G$ is not entirely free and been restrained by the parameters in discriminator $D$. Hence the possible overfitting of GAN, such as in the scene of generating numerical data, has not attracted enough attention and been studied thoroughly. Lastly unless been concretely discussed, adjusting the DoF of $G$ randomly could not improve the augmentations. As been shown in Appendix table 1, DoF of $G$ has been changed either positively or negatively, while $D$ maintains, the AUC of the RFC classifier after data augmentation has not been improved.

**Table 2** Comparision of the Dimension and DoF in Datasets and GANs

| **Dataset dimension** | *Dataset[1]*:14 | *Dataset[2]*:20 | *Dataset[3]*:8 | *Dataset[4]*:22 |
|---|---|---|---|---|
| **Model parameter** | GAN:6k-10k | GAN-DAE:7k-20k | WGAN:6k-10k | WGAN-GP:6k-12k |

To study the possible overfitting that not received enough attention, we first theoretically study the GAN by analyzing the Directed Graphical Model (DGM) [10]. Then we design a restraint function as an extra flow directed *G* from *D* in DGM. Such restraint is desired to be both independent of the current loss function and quantifiable by practical functions. We expect such restraint could suppress overfitting and results in a high AUC in the augmentation dataset. A Statically Restrained GAN (SRGAN) and Dynamically Restrained GAN (DRGAN) are designed for practicing this theoretical model. The SRGAN predefines a serious of particular network topology pairs and quantified pairs by a Similarity of the Restraint (*SR*) function, which concretely interprets the theoretical restraint function. While, in DRGAN, an adjusting dropout function is designed based on the KMMD[11].

In proving the feasibility of theory and our design's effectiveness, we carry out the experiments on four standard numerical imbalanced datasets and five classifiers. Compared with the two baselines, as original data and SMOTE[9] generated data, and three competitors of different types of GANs, SRGAN and DRGAN together produce the best-1 augmentation result of 19 from 20; while in measuring the best-2 augmentation methods, the SR and DR methods lead in 14 of 20 groups of the top-2. In the ablation experiments, we find in the 19 best-1 augmentation results, the DRGAN overperform SRGAN in 13:6, while SRGAN could interpret the restrain more concretely with a proportional relationship with the *SR* and final matric AUC.

This paper's innovation: we address possible overfitting in the numerical data augmentation by GAN and propose a theoretical restraint for *D* to *G* to suppress overfitting. Such restraint is independent of the loss functions and quantifiable, hence study the restraints could lead to a deeper insight into the adversarial strategy. Practically, the static and dynamic restraint enrich design ideas for improving the quality of *G* generated data.

The remainder of the paper is organized as follows. We introduce an overview of previous related works on GAN and data augmentation in Section 2. We provide the proposed statically and dynamically restrained GAN and the analysis through DGM in detail in Section 3. We show the further improved performance of statically and dynamically restrained GANs than original data, SMOTE, and other GANs using five classifiers on four representative datasets in Section 4. Finally, Section 5 presents the conclusions and outlines possible directions for future research.

## 2  Related Works

Currently, many GAN models [14, 15] have made significant progress in generating images and improving the accuracy of classifiers, some of which can produce almost indistinguishable images from human visual inspection. GAN is a deep learning model for game learning between generator *G* and discriminator *D*. *G* is a network model that generates data by receiving a random noise z and generating data from this noise, while *D* is a discriminant network model used to distinguish the data generated by *G* from the real data samples in the dataset. In the training process, G's goal is to use the generated data to confuse *D*, while D's goal is to distinguish the generated data from the real data, thus constituting a dynamic game process between *G* and *D*. However, most of the work is restored to other specific dataset scenarios, such as numerical data. Because GAN models have problems such as model collapse and loss function non-convergence of generators, and *G* over-fitting, resulting in poor data results [1,6]. Many works of literature have done much research on the improvement of the GAN model. In the course of the study, the following aspects are discussed,

1) Improve the structure of GAN model, such as training the interaction between multiple discriminator *D* and multiple generator *G*, such as DualGAN [16] model consisting of two *D* and two *G*. In recent years, GAN models have been used to generate value data to enhance credit card fraud detection through data argumentation [12,17,18], and the performance of the classifier in fraud detection, using Denoising Auto encoder (DAE) to learn the complicated probability relationship between input features effectively, and using antagonistic learning method, to establish a minimum-maximum game between discriminator and generator to discriminate positive and negative samples accurately[13]. Larsen et al. [19] proposed an automatic encoder that uses learned representations to measure similarity in data space better. By combining Variational Auto-encoder (VAE) with GAN, GAN discriminators' learning characteristics can be represented as the basis for the goal of VAE reconstruction.

2) GAN's loss function is improved, such as WGAN [6], LSGAN [3]. WGAN solves model collapse and GAN training instability very well and ensures the diversity of generated data [6].

3) Additional information is added to the GAN model to generate data of the target class such as conditional information c input to the network, CGAN [20] to control the conditional output mode of the network, and InfoGAN [21] to generate samples of the same class by maximizing mutual information (c, c').

Evaluating and comparing the generated data is a challenge for GAN. This is partly due to the lack of a precise likelihood measure [22], common in comparable probability models [23, 24]. Therefore, many previously generated image data are evaluated using subjective visual qualities. It is difficult to accurately judge the quality of GAN generated sample images by subjective evaluation. Inception Score (IS) [25] and Fr Chet Inception Distance (FID) [26] are the most widely used indicators in sample-based quantitative assessment measures. However, these two indicators can only be used to evaluate image datasets, and we need to evaluate the generated numerical data. So we measure [11] by generating numeric data based on Kernel Maximum Mean Discrepancy (*Kernel MMD*). *Kernel MMD* is characterized by low computation cost, excellent performance, and computing structured data.

## 3 Proposed Method

### 3.1 A Theoretical Restrained Structural in DGM for GAN

To explore the mechanism of GANs, a directed graphical model (DGM) [10] is setup to analyze the process of GAN. Firstly, the parameters set of *G* and *D* are defined as $\omega_G$ and $\omega_D$. The general loss function is separated into GAN's loss function in GAN is defined as $loss_G$ and loss function of *D* as $loss_D$. Different GANs models may specify different loss functions, such as GAN [1] and WGAN [6]. The loss function of GAN and WGAN is shown as follow:

$$\min_\theta \max_\omega E_{x \sim P_r}[log D_\omega(x)] - E_{z \sim P_z}[log D_\omega(g_\theta(z))] \quad (1)$$

$$\min_\theta \max_\omega E_{x \sim P_r}[f_\omega(x)] - E_{z \sim P_z}[f_\omega(g_\theta(z))] \quad (2)$$

In DGM, functions for either GAN are simplified to $loss_G$ and $loss_D$. According to study of the network [30], the loss function is variable on a complex functional space and the parameters set $\omega_G$, $\omega_D$. Mathematically, we consider $\omega_G$, $\omega_D$ are represented as variables in Hilbert space, and $loss_G$ is a variable defined in a functional space. Hence, we can use DGM to describe G's training process in GAN, with nodes of $\omega_G$, $\omega_D$ and $loss_G$, as demonstrated in Fig. 1.(a). Subsequently, learning $G$ is to minimize the loss function $f_{loss_G}$, equal to maximize the *posterior* probability of $\omega_D$, given $\omega_D, loss_G, loss_D$, as follow

$$\omega_G = argmin f_{loss_G} = arg \max_{\omega_G} p(\omega_G | \omega_D, loss_G, loss_D) \qquad (3)$$

In DGM, a factorization is operated and hence there is,

$$f(\omega_G | \omega_D, loss_G, loss_D) = f(\omega_G | \omega_D, loss_G) \qquad (4)$$

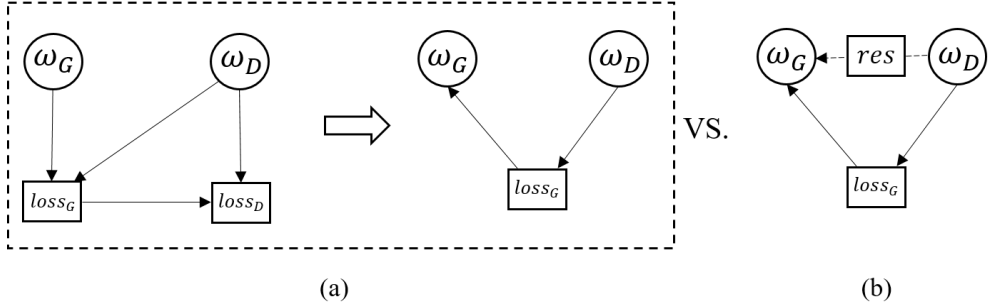

(a)            (b)

**Fig. 1.** (a) GAN structure diagram based on DGM analysis (b) Restrained GAN structure diagram based on DGM analysis. The circle nodes are the parameter set for D and G which are variable in a Hilbert Space [28]; while the rectangle nodes are the variables in a functional space.

Such factorization results in a simplified DGM Fig. 1. (b) that we can calculate $\omega_G$ given $\omega_D$ and a loss function $loss_G$, since $loss_G$ is separated from the general $loss_D$. Now assume an extra function as a functional node is added directing from $D$ to $G$, as a function $f_{res}$, we can see, in In Fig. 1. (c), the new flow direction, $\omega_G \xleftarrow{res} \omega_D$ is added to the DGM. By introducing restraints from $D$ and $G$ during GAN training. The function $f$ of GAN training process of $G$ changes to

$$f(\omega_G | \omega_D, f_{res}, loss_G, loss_D) = f(\omega_G | \omega_D, f_{res}, loss_G) \qquad (5)$$

If the restraint function $f_{res}$ depends on the loss function in the functional space, as $f_{res} \perp loss_G$, we could have:

$$f(\omega_G | \omega_D, f_{res}, loss_G) = f(\omega_G | loss_G, \omega_D) f(\omega_G | f_{res}, \omega_D) \qquad (6)$$

From equation 6, we can draw such conclusion that, in training $\omega_G$, two independent factors are working at the same time, where one is the conventional training of $G$, given $loss_G, \omega_D$, and the other is from the restraint we design and desire to suppress the possible overfitting. Under the independence condition, we can practice a measurable and interpretable restraint function to optimize $G$ without affecting the conventional GAN training. Such a function could obtain information for $D$ and enhance $G$ in training as a restraint. In the light of GAN models, $D$ is a strong classifier since its task is simply to identify the generated and original sample, while $G$ is a much weaker generator that maps low dimensional noises to high dimensional samples; subsequently, a restraint from $D$ to $G$ could lower $G$'s DoF and hence suppresses the possible overfitting.

To implement the theory, we need to design a $f_{res}$ satisfying two request, 1) $f_{res}$ should be independent of $loss_G$; 2) $f_{res}$ should be represented measurably and interpretably instead of a training process or structure, only. In the following two sections, we present two models with different ideas in practicing $f_{\text{res}}$. One is named statically restrained GAN, where the $f_{\text{res}}$ is regarded as an implicit function of the topologies for the *G-D* pairs and measured by a *SR* metric. The other one is called dynamically restrained GAN that adjusts the generator structure based on the $f_{\text{res}}$ calculated from a KMMD-based dropout rate in training.

### 3.2 Statically Restrained GAN

In this section, we present a Statically Restrained Structural Solutions, in the idea to implicitly implement $f_{res}$ as pairs of predefined network topologies of *G* and *D*. The predefined topologies fit a particular pattern of networks in layers and nodes. Then we quantify the restraint by the *SR*. Higher *SR* indicating a higher similarity between networks, and subsequently a stronger restraint from *D* to *G*.

Practically，We design several SRGAN structures between *G* and *D*, naming *isomorphic, axisymmetric, self-symmetric*, and *axisymmetric* and *self-symmetric G-D* pairs, shown in Fig. 2. Here the isomorphic structure is defined as that the two networks have the same number of layers, each layer has the same number of nodes, and every two neighboring layers have the same connection. Similar to the definition of isomorphic structure, the axisymmetric structure is defined as an axisymmetric network structure; the self-symmetric structure is defined as the symmetrical network structure. Besides, the axisymmetric and self-symmetric structure is defined as both symmetrical and axisymmetric network structure.

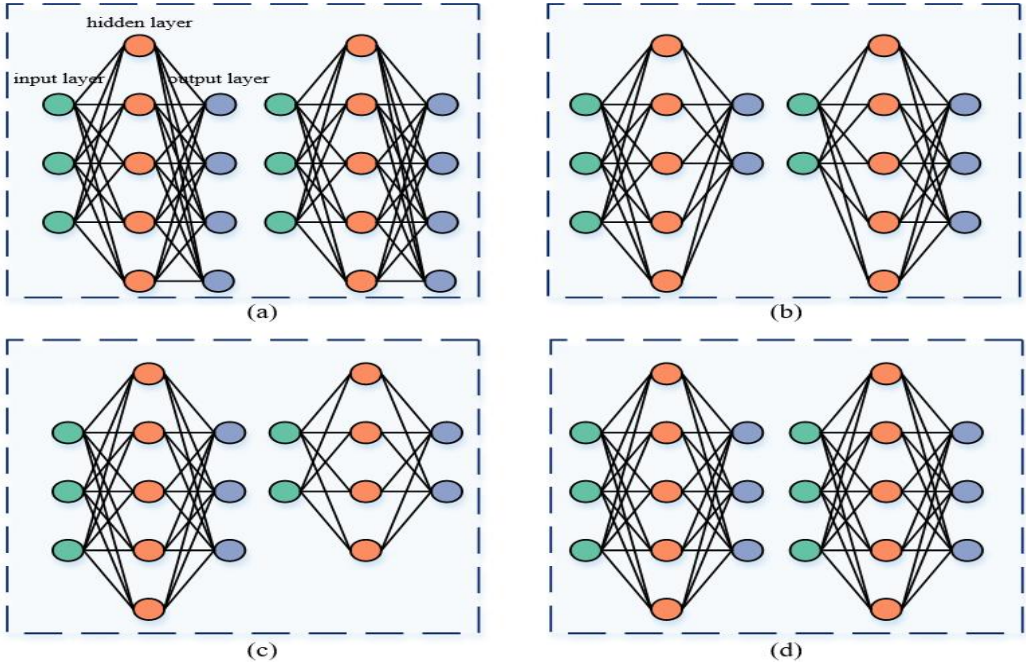

**Fig. 2.** The network structure of *G* and *D* in (a) isomorphic structure, (b) axisymmetric structure, (c) self-symmetric structure and (d) axisymmetric and self-symmetric structure

The implicit topology restraint is measured by a quantitative matric *SR*, motivated by Tian et al. [25], as follows:

(7)

$$f_{\text{res}} = SR = mean(\sum_{i=0}^{n} max\rho_{jj°})$$

Here $f_j = [f_j(x_1),...,f_j(x_k)]$ and $f_{j°} = [f_{j°}(x_1°),...,f_{j°}(x_k°)]$ $\rho_{jj°} = \tilde{f}_j^T \tilde{f}_{j°} \in [-1,1]$, where $\tilde{f}_j = [f_j - mean(f_j)/std(f_j)]$. Besides, where $f_j$ and $\tilde{f}_{j°}$ are the weight of each node in the neural network of D and G, respectively. Furthermore, we normalize $f_j$ and $f_{j°}$ to get $\tilde{f}_j$ and $\tilde{f}_{j°}$. The correlation coefficient $\rho_{jj°}$ is obtained by the inner product of $\tilde{f}_j$ and $\tilde{f}_{j°}$ of D and G.

SR indicates how similarly the two layers in two networks respond with the same inputs, i.e. the similarity of two networks. According to this definition, SR concretely interprets the restraint in structure level. If G and D can respond the same input similarly with high SR, then G is deemed to be adequately restrained by D. SR is an effective interpreter for $f_{\text{res}}$, despite that the two networks undertake different tasks as a classifier ranging in {0,1} and a generator ranging in high dimensional sample spaces. Experiments in Section 4.3 demonstrate a proportional relationship between SR and the generation, measured by AUC, proving the innovativeness and feasibility of SR as an adequate representation of $f_{\text{res}}$.

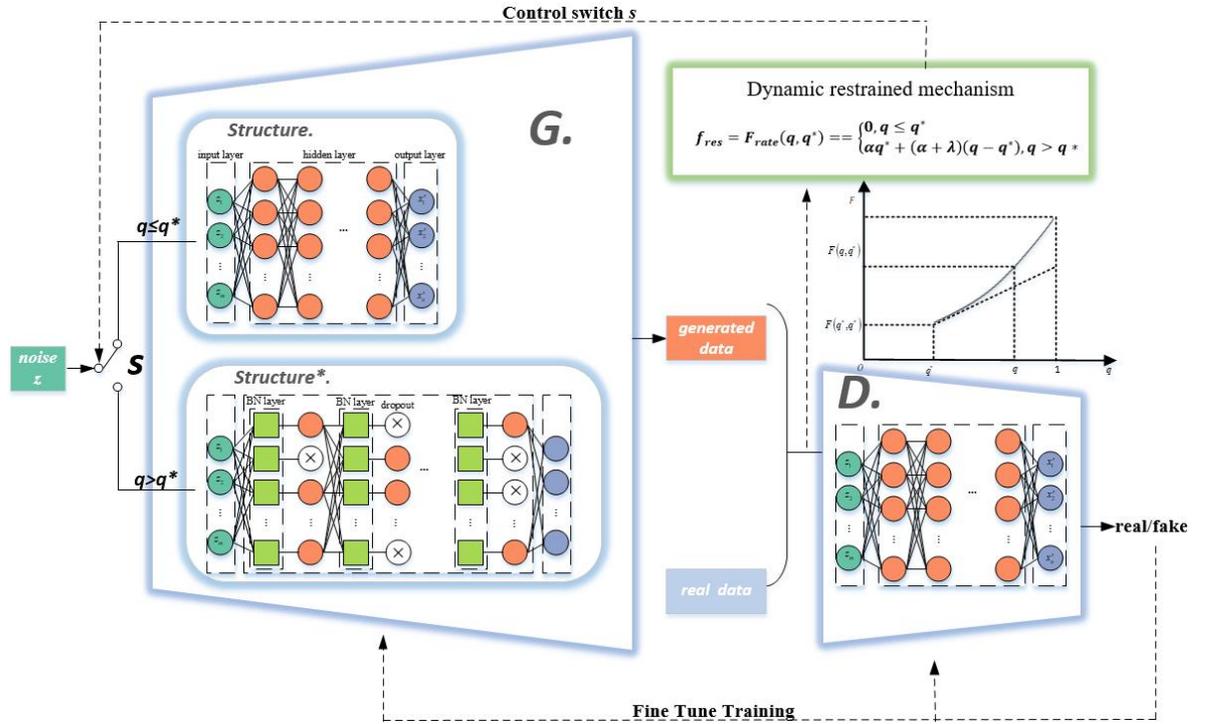

**Fig. 3.** Dynamically restrained GAN training strategy of dynamic restrained mechanism

**3.3 Dynamically Restrained Structural Solutions**

In SRGAN, we employ the restraint in an implicit way, but with a concretely interpretable metric that explains how G is passively limited the strong classifier D. Instead of using this predefined and fixed network pairs that are empirically designed and ignoring the information from the dataset, we present another way to realize restraint is to adjust the G's structure dynamically based on $f_{res}$ in this section.

Since the concept of restraint is proposed to suppress overfitting, it is quite natural to introduce dropout as a dynamic adjustment of G. Here, we design a piecewise nonlinear dropout rate function as $f_{res}$ based on the third-party metric *Kernal MMD*(KMMD) [11] which is independent to the loss function *G* and measured the quality of the generation samples. Note that many metrics, such as IS [25], FID [26], KMMD, as soon as they are presented independently of the loss function, could be used to design $f_{res}$. Here we employ KMMD because it is the fastest metric, characterized by low computation cost and good performance so that it can be calculated in every iteration efficiently. Meanwhile, despite the independence, these metrics carry out the same task of *D*, evaluating the quality of the generation, hence they could be considered an implicit restraint from *D* to *G*.

The dropout ratio is designed using the punishment mechanism function using the evaluation index of the generated data from each training.

$$f_{res} = F_{rate}(q, q^*) == \begin{cases} 0, q \leq q^* \\ \alpha q^* + (\alpha + \lambda)(q - q^*), q > q* \end{cases} \tag{8}$$

where $q$ and $q^*$ are the min-max normalized indices from KMMD, $q$ is the index form the current iteration, while $q^*$ is the current optimized index in training. Noted that a lower $q$ is expected for a generated dataset in better quality, and if $q$ increases during the training, then possible overfitting is deemed, and hence a dropout is carried out, based on the linear base dropout rate of $\alpha q$ and a nonlinear penalty of $(\alpha + \lambda)(q - q^*)$. Generally, $\lambda$ indicates the strength of the restraint, however, it could be quite hard to illustrate the restraint strength concretely. According to the dropout definition [27], the dropout is not proportionally related to the neural networks' performance, so that to the restraint effectiveness. Experiments in Section 4.3(c) demonstrate that there is a peak in Fig.5. Despite the weak interpretability, experiments in Section 4.3(a) show that a DRGAN generally overperforms SRGAN, where both of them surpass the other GANs and baselines.

## 4 Experiments and Discussion

In this section, we select four different binary numerical datasets from the University of California Irvine (UCI) machine learning repository to model different GANs using static and dynamic strategies to adjust the network structure of GANs. The numeric data generated by *G* of the trained model is augmented, and the ability to improve the AUC of five classifiers is experimentally studied. Simultaneously, the configuration and environment of the experiment are introduced, and the experiment design and result analysis are described and discussed.

### 4.1 Experiments

In the experiments, we used four different imbalanced numerical datasets from the UCI Machine Learning Repository [8], namely the Australian Credit Approval dataset, the German Credit dataset, the Pima Indians Diabetes dataset, and the SPECT Heart dataset. The basic overview of the dataset is shown in Table 3.

**Table 3** Four Numerical Datasets Used in the Experiments

| Name of Dataset | No. of samples | No. of attributes |
|---|---|---|
| Australian Credit Approval dataset | 690 | 14 |
| German Credit dataset | 1000 | 20 |
| Pima Indians Diabetes dataset | 768 | 8 |
| SPECT Heart dataset | 267 | 22 |

For four datasets, we selected Artificial Neural Network (ANN), Support Vector Machine (SVM), K-Nearest Neighbor (KNN), Random Forest Classifier (RFC), and Gradient Boosting Classifier (GBC). Five classifiers test the performance of each GAN trained with static and dynamic adjustment network structure strategies recommended by Scikit-learn [7]. Hence 20 groups of experiments are carried out where 20 equals 4 datasets times 5 classifiers. SRGAN and DRGAN are compared with traditional training GAN, GAN-DAE, and WGAN. Because the accuracy does not apply to imbalanced data, we select AUC (Area Under Curve) as the classifier's evaluation index. AUC is a widely-accepted evaluation index to measure the two-classification model's quality in imbalanced data augmentation [29, 32, 33]. To reduce randomness on the generation models, 10-fold (K=10) cross-validation experiment is carried out in the experiments.

**4.2 General Performance on Restrained GAN**

The section generally demonstrates the outperformance of SRGAN (AWGAN, SWGAN, ASWGAN, IWGAN) and DRGAN (WGAN*) in AUC in 20 groups of experiments. Fig. 4 illustrates the comparison of multiple SRGAN (in blues) and DRGAN verse the baselines of Original data without data augmentation, with SMOTE augmentation and GAN (GAN, GAN-DAE, WGAN) augmentations. Generally speaking, AUC in both SRGAN and DRGAN concretely exceeds the baselines in most datasets and classifiers. More specifically, among the four UCI datasets and five classifiers, German Credit datasets are the most sensitive to classifiers, and data augmentation method DRGAN is the best for AUC promotion. For different classifiers, the DRGAN model has the best AUC promotion performance on all datasets in ANN and GBC classifiers, while the KNN classifier has the best AUC enhancement performance in addition to the dynamic restrained WGAN in German Credit dataset. For SVM classifiers, the AUC is improved by 7.1% over the original data except for the data augmentation effect in the SPECT Heart dataset. In the other three datasets, various data augmentation algorithms have no significant effect on AUC promotion. Lastly, in figure 4 we can see that WGAN, GAN-DAE, GAN, and SMOTE data augmentation methods are affected by different datasets and classifiers, resulting in unstable data augmentation. Hence, we summarize the general performance in the best and top-2 ranking table.

In table 4, we merge various SRGANs, including AWGAN, ASWGAN，SWGAN, and IWGAN, as a general type SRGAN, to simplify the entity list. We can see that among all 20 experiments, SRGAN or DRGAN won the first place in 19 experiments. Only on the Pima Indians Diabetes dataset and under the KNN classifier, SMOTE augmented data reaches the highest AUC. In terms of statistics, we have counted the augmentation effects of top-2 and hope that SRGAN and DRGAN can take the top two. In this case, we see that among the 19 groups of experiments where the existing restrained GAN won the first place, there are 14 groups of SRGAN and DRGAN win the first two positions simultaneously. Whereas in the other 5 groups of experiments, another argumentation algorithm (original data without argumentation) won the second place.

In-depth analysis of this shows that of the 6 unsatisfactory cases, 3 of them all occur under the Pima Indians Diabetes dataset. The non-restrained GAN algorithm captures 1 group of the first place and 2 groups of the second place. This is due to the Pima Indians Diabetes dataset's low dimension, as in 8 dimensions. This aspect confirms our conjecture that under low-dimensional data sets, the GAN-based argumentation algorithm is very inclined to overfitting, resulting in poor results. This proves the necessity of restraint in the GAN algorithm from another view. At the same time, we are also aware that the existing two restraint designs may not have enough strength. In the future, we may need to design further a GAN algorithm that can achieve more substantial constraints to generate low-dimensional data.

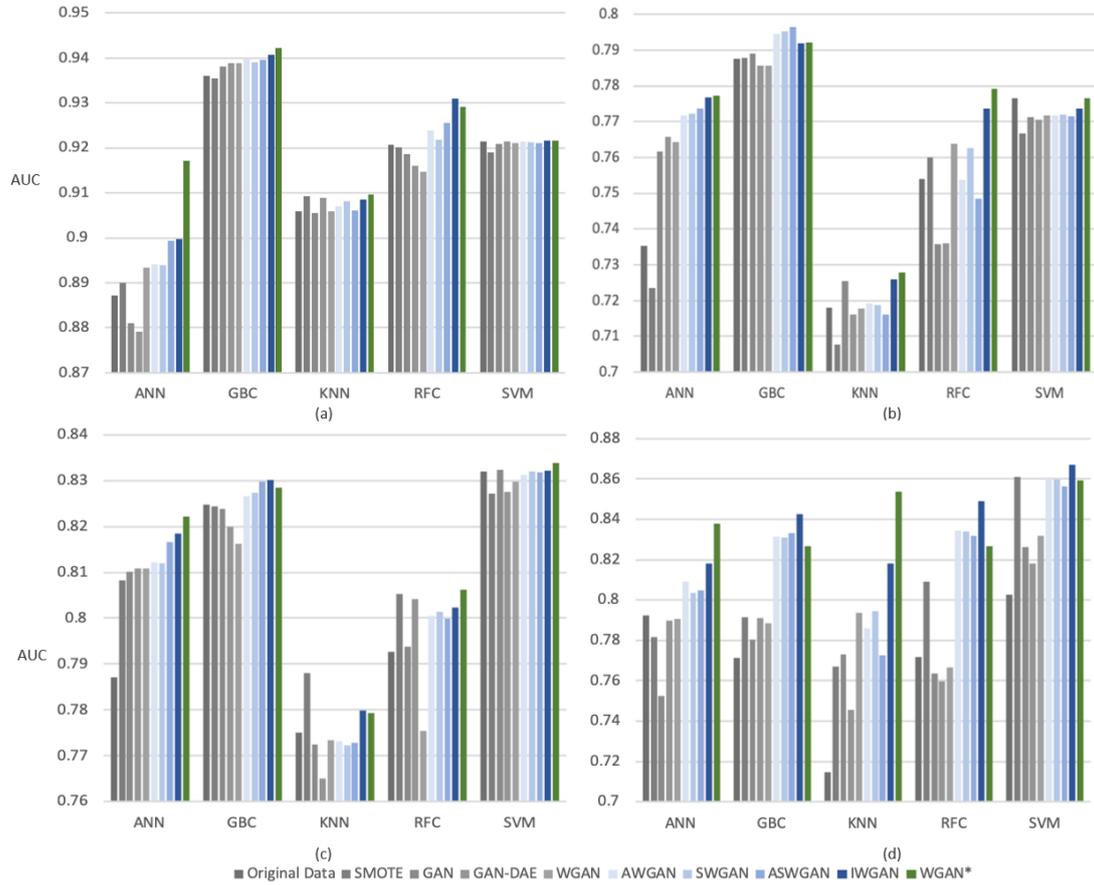

**Fig. 4.** Nine data augmentation methods of AUC for five classifiers in (a) Australian Credit Approval dataset (b) German Credit dataset (c) Pima Indians Diabetes dataset and (d) SPECT Heart dataset

Table 4 Nine data augmentation methods of AUC for five classifiers on four datasets in UCI

|  | ANN | GBC | KNN | RFC | SVM |
| --- | --- | --- | --- | --- | --- |
| Australian Credit Approval dataset (Dimension:20) | | | | | |
| Best | DRGAN | DRGAN | DRGAN | SRGAN | DRGAN |
| Second Best | SRGAN | SRGAN | *SMOTE* | DRGAN | SRGAN |
| German Credit dataset (Dimension: 14) | | | | | |
| Best | DRGAN | SRGAN | DRGAN | DRGAN | DRGAN |
| Second Best | SRGAN | DRGAN | SRGAN | SRGAN | *Original Data* |
| Pima Indians Diabetes dataset (Dimension: 8) | | | | | |
| Best | DRGAN | SRGAN | *SMOTE* | DRGAN | DRGAN |
| Second Best | SRGAN | DRGAN | SRGAN | *SMOTE* | *GAN* |
| SPECT Heart dataset (Dimension: 22) | | | | | |
| Best | DRGAN | SRGAN | DRGAN | SRGAN | SRGAN |
| Second Best | SRGAN | DRGAN | SRGAN | DRGAN | *SMOTE* |

**4.3 Discussion on Restrained GAN**

In this section, we try to evaluate and discuss a series of performances in RGAN.

**a) Static Restrained GAN V.S. Dynamic Restraind GAN**

In this section, we compare the performance differences between the two restraints. In 20 experiments, We find that 7 groups perform better with SRGAN and the remaining 13 groups perform better with DRGAN. Generally, the restraints of DRGAN are better than SRGAN. It is noted that for the performance of two classifiers ANN and SVM, DRGAN in ANN is overall better than SRGAN. We suspect this is because DRGAN itself is optimized by adjusting the network structure dynamically, which is more targets to enhance such a neural network classifier ANN. In Section 4.2, we find that in 6/20 experiments, SRGAN and DRGAN failed to achieve the top-2 simultaneously, and we speculate that 3/6 groups all appear in a particularly low-dimensional as Pima Indians Diabetes dataset. Another thing worth noting is the SVM classifier, such a classic discriminant model, the remaining 3/6 groups all appear under the SVM classifier in the comparative discussion. Besides, the SVM classifier presents an extreme state for SRGAN and DRGAN, one take the first place, the other one could not even get into the top-2. We guess that this is the data generated by the two restrained algorithms, which is extreme in the hyperplane mapped by the SVM kernel. When one is superior, the other is relatively poor. This indicates our future direction for the influence of restraints on the network distribution changes in the GAN functional space.

**b) Interpretation of Static Restrained GAN and SR**

Given that this paper's core application work is to practice a function like $f_{\text{res}}$, in this section, we try to understand the relationship between the static restraint matric *SR* and strength of $f_{\text{res}}$, which be measured implicitly by the AUC.

To expend this idea of static restraint, we set various topologies pair of *D* and *G*, but not limit them to IWGAN, SWGAN, etc. Each topology concludes that the *SR* and final AUC of the generated data on different classifiers is calculated in which we expect to see a proportional relationship. Such a proportion indicates that *SR* could positively reflect the strength of restraint, subsequently leading to a suppressed overfitting in generation and results in a higher AUC. From Fig. 5, we observed a positive correlation between AUC and *SR* under static restraints on four datasets.

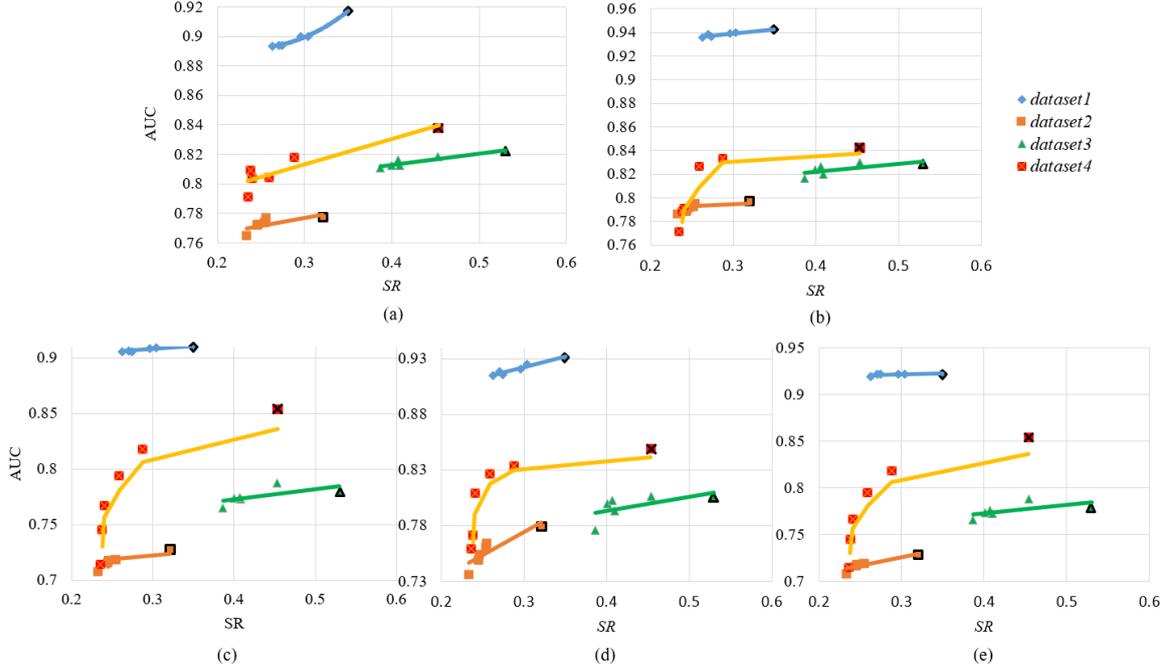

**Fig. 5.** Static restrained GAN methods of SR and AUC for four datasets in (a) ANN (b) GBC (c) KNN (d) RFC and (e) SVM.

Each dataset performs variously on different classifiers. In general, the classification effect of *Dataset[1]* (Australian Credit Approval dataset) is the most obvious, while the *Dataset[2]* (German Credit dataset) effect is relatively weak. Overall, we can see that on different classifiers, for *Dataset[1]*, *Dataset[3]* (Pima Indians Diabetes dataset) and *Dataset[4]* (SPECT Heart dataset), the relationship between *SR* and AUC is relatively linear, which proves that when we design topology pairs to improve *SR*, the restraint (potentially) linearly enhanced, thereby exhibits a linearly enhanced AUC. For *Dataset[2]*, when the *SR* is increased, it will firstly bring a dramatically increasing and then close to flat. This shows that for this dataset, a better topology pair with higher *SR* will first have a more significant impact, and then become relatively stable.

In summary, we can find that *SR* can directly reflect AUC to interpret potential $f_{res}$ better. On the other hand, the effect of *SR* or static restraint may be relatively distinct for different datasets. In some datasets, specific network restraints with higher *SR* may be better than other algorithms in stages. Further in-depth research is subject to follow-up investment.

**c) Analysis of Dynamically Restrained GAN and $F_{rate}$**

In this paragraph, we try to understand the relationship between dropout rate $F_{rate}$ and $f_{res}$ in DRGAN. In the past paper[34], it can be seen that a reasonable preferred dropout rate is 0.2 to 0.6. We believe that the dropout effect is the most obvious in this interval, and further restraint is stronger. Based on this, we list the AUC changes brought about when λ changes.

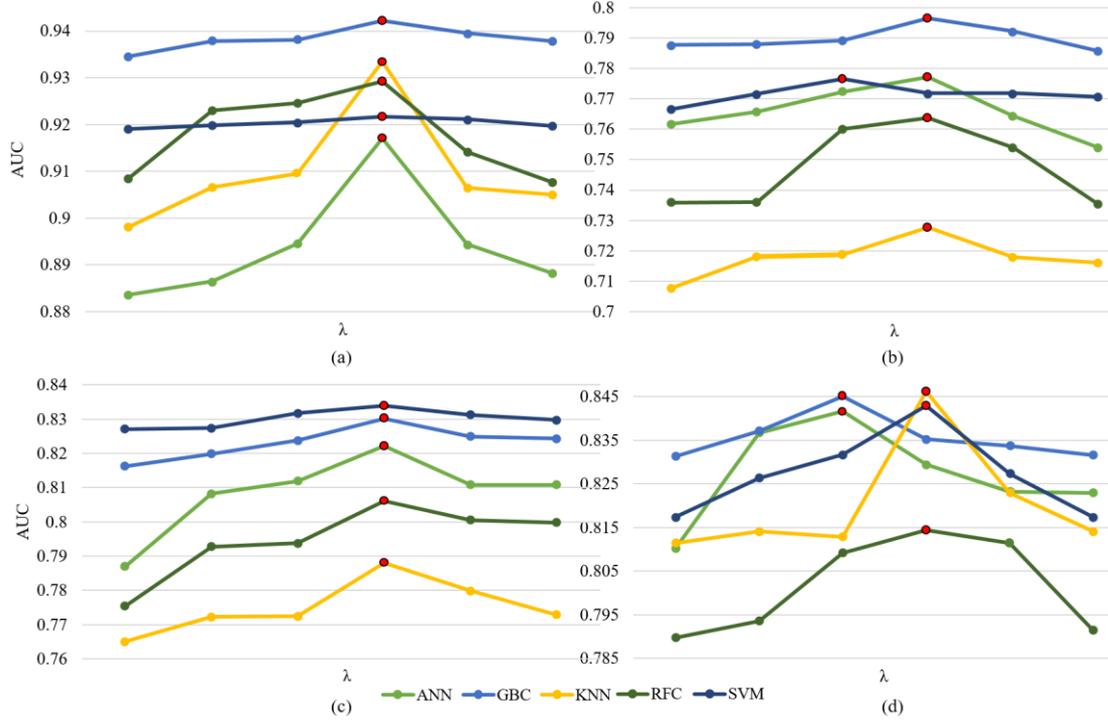

**Fig. 6.** Dynamic restraint strength λ of AUC for five classifiers in (a) Australian Credit Approval dataset (b) German Credit dataset (c) Pima Indians Diabetes dataset and (d) SPECT Heart dataset

Fig. 6. shows that in each dataset, the changes in AUC under different classifiers are consistent with the varieties in λ. When λ increases, AUC first increases and then decreases. According to formula 8, $F_{rate}$ and λ are linearly related with each other, and the AUC peak appears at a certain point of $F_{rate}$ simultaneously, which is consistent with our conclusion that a reasonable preference for the dropout rate is 0.2 to 0.6.

The peak of λ appears to be relatively fixed from the multiple datasets, which means that the preferred $F_{rate}$ situation is relatively stable. From the perspective of multiple classifiers, the sensitivity of different classifiers to λ changes is different. In contrast, ANN improves the most, which matches our previous conclusion that ANN is most affected by the network results in restraint GAN. In comparison, SVM and RFC are not sensitive to the changing trend of λ, to a certain extent showing that these two classifiers are not very practical for GAN changes under dynamic restraints, which is similar to the previous conclusions. This leads us to observe and study the sensitivity of different classifiers to different restraints, so as to study better constraint methods in the future.

## 5  Conclusion and Future Extensions

In this paper, we have addressed a possible overfitting problem in numerical data generation. Via the DGM study, a theoretical Restrained model is proposed with an independence function $f_{\text{res}}$. Further we propose two practical designing of $f_{\text{res}}$, as Statically Restrained GAN (SRGAN) and Dynamically Restrained GAN (DRGAN). On these model, the $f_{\text{res}}$ are interpreted as metric Similarity of the Restraint (SR) and dynamical dropout rate $F_{rate}$. Experimental results show that our overfitting problem can be solved by adding $f_{\text{res}}$ restraint. In the experiment of 5 classifiers and 4 datasets, 19 groups of restrained GAN win the first place, and 14 groups of SRGAN and DRGAN win the top-2 places at the same time.

This paper focuses on the overfitting problem of GAN in low dimensional data generation, and proposes a theoretical solution. On this basis, there may be many ways to implement this restraint, more effective restraint methods are expected to be proposed and implemented in future research.

## ACKNOWLEDGMENTS

This work was supported by National Key Research and Development Program of China under the grant number (2018hjyzkfkt-002), Science Challenge Project of China (TZZT2019-B1.1), China NSF under the grant number (11771440) and the Tianjin Natural Science Foundations (17JCYBJC23000).

## APPENDIX

Table 1 Failures of data augmentation by GANs (Evaluated by classifier RFC and the top-2 are bold)

| Techniques | Dateset[1] | Dataset[2] | Dataset[3] | Dataset[4] |
|---|---|---|---|---|
|  | | AUC | | |
| Baseline[1]-SMOTE [9] | **0.9201** | **0.7600** | **0.8052** | **0.8092** |
| Baseline[2]-Original data | **0.9207** | 0.7540 | 0.7927 | **0.7716** |
| GAN ↓ | 0.9166 | 0.7328 | 0.7883 | 0.7622 |
| GAN [1] | 0.9186 | 0.7359 | 0.7937 | 0.7634 |
| GAN ↑ | 0.9193 | 0.7347 | 0.7860 | 0.7608 |
| GAN-DAE ↓ | 0.9168 | 0.7389 | **0.8054** | 0.7487 |
| GAN-DAE [13] | 0.9161 | 0.7360 | 0.8042 | 0.7598 |
| GAN-DAE ↑ | 0.9179 | 0.7323 | 0.7998 | 0.7523 |
| WGAN ↓ | 0.9149 | 0.7325 | 0.7883 | 0.7466 |
| WGAN [6] | 0.9147 | **0.7638** | 0.7755 | 0.7665 |
| WGAN ↑ | 0.9114 | 0.7465 | 0.7724 | 0.7511 |

Note: On the four datasets, all GAN-based methods generate worse argument data, evaluated by the classifier RFC, than the **Baseline[1]-SMOTE** and **Baseline[2]-Original data**. ↓ and ↑ represents the parameters of reducing and increasing *G* respectively. (*Dataset[1]*: Australian Credit Approval dataset, *Dataset2*: German Credit dataset, *Dataset[3]*: Pima Indians Diabetes dataset, *Dataset[4]*: SPECT Heart dataset)

Table 2 Nine data augmentation methods of AUC for five classifiers in Australian Credit Approval dataset

|  | ANN | GBC | KNN | RFC | SVM |
|---|---|---|---|---|---|
| Original Data | 0.88726 | 0.93607 | 0.90593 | 0.92065 | 0.92136 |
| SMOTE | 0.88996 | 0.93548 | 0.90935 | 0.9201 | 0.91901 |
| GAN | 0.88106 | 0.93816 | 0.90554 | 0.91859 | 0.92093 |
| GAN-DAE | 0.8792 | 0.93879 | 0.90891 | 0.91609 | 0.92137 |
| WGAN | 0.89334 | 0.93888 | 0.90598 | 0.9147 | 0.92101 |
| AWGAN | 0.89406 | 0.94 | 0.90703 | 0.92392 | 0.92145 |
| SWGAN | 0.89401 | 0.93895 | 0.90812 | 0.92174 | 0.92131 |
| ASWGAN | 0.8994 | 0.93958 | 0.9061 | 0.92563 | 0.9211 |
| IWGAN | 0.89975 | 0.94075 | 0.90853 | 0.93095 | 0.92155 |
| WGAN* | 0.91711 | 0.94214 | 0.90956 | 0.92918 | 0.92162 |

Table 3 Nine data augmentation methods of AUC for five classifiers in German Credit dataset

|  | ANN | GBC | KNN | RFC | SVM |
|---|---|---|---|---|---|
| Original Data | 0.73523 | 0.78764 | 0.7181 | 0.75404 | 0.77651 |
| SMOTE | 0.72363 | 0.7879 | 0.70775 | 0.76001 | 0.76665 |
| GAN | 0.76167 | 0.78912 | 0.72558 | 0.73587 | 0.77122 |
| GAN-DAE | 0.76577 | 0.78566 | 0.71617 | 0.73601 | 0.77057 |
| WGAN | 0.76424 | 0.78571 | 0.7179 | 0.76375 | 0.77183 |
| AWGAN | 0.7717 | 0.79455 | 0.71915 | 0.75388 | 0.77179 |
| SWGAN | 0.7723 | 0.79531 | 0.71878 | 0.76272 | 0.77208 |
| ASWGAN | 0.77359 | 0.79642 | 0.71612 | 0.74861 | 0.7716 |
| IWGAN | 0.77676 | 0.79192 | 0.72596 | 0.7736 | 0.7736 |
| WGAN* | 0.77718 | 0.79214 | 0.72776 | 0.77918 | 0.77662 |

Table 4 Nine data augmentation methods of AUC for five classifiers in Pima Indians Diabetes dataset

|  | ANN | GBC | KNN | RFC | SVM |
|---|---|---|---|---|---|
| Original Data | 0.78701 | 0.82483 | 0.77507 | 0.79273 | 0.83204 |
| SMOTE | 0.80824 | 0.82437 | 0.78808 | 0.80523 | 0.82715 |
| GAN | 0.81004 | 0.82373 | 0.77246 | 0.79374 | 0.83237 |
| GAN-DAE | 0.81075 | 0.81992 | 0.76506 | 0.8042 | 0.82745 |
| WGAN | 0.81085 | 0.81626 | 0.77341 | 0.77547 | 0.82971 |
| AWGAN | 0.81206 | 0.82658 | 0.77322 | 0.80054 | 0.83129 |
| SWGAN | 0.81198 | 0.82736 | 0.77227 | 0.80136 | 0.83205 |
| ASWGAN | 0.81652 | 0.82969 | 0.77291 | 0.79981 | 0.83173 |
| IWGAN | 0.81848 | 0.83021 | 0.77985 | 0.80232 | 0.83212 |
| WGAN* | 0.82217 | 0.82837 | 0.77922 | 0.80618 | 0.83392 |

Table 5 Nine data augmentation methods of AUC for five classifiers in SPECT Heart dataset

|  | ANN | GBC | KNN | RFC | SVM |
|---|---|---|---|---|---|
| Original Data | 0.79219 | 0.77133 | 0.71449 | 0.77155 | 0.80256 |
| SMOTE | 0.78165 | 0.79127 | 0.76682 | 0.80924 | 0.8608 |
| GAN | 0.75238 | 0.78011 | 0.77306 | 0.76341 | 0.82638 |
| GAN-DAE | 0.78963 | 0.79086 | 0.74561 | 0.75979 | 0.81795 |
| WGAN | 0.79059 | 0.78837 | 0.79362 | 0.76648 | 0.83168 |
| AWGAN | 0.80928 | 0.83128 | 0.7858 | 0.83452 | 0.8604 |
| SWGAN | 0.80339 | 0.83099 | 0.7943 | 0.83387 | 0.8596 |
| ASWGAN | 0.8046 | 0.8331 | 0.77256 | 0.83202 | 0.8563 |
| IWGAN | 0.81806 | 0.84246 | 0.81822 | 0.84918 | 0.867 |
| WGAN* | 0.83803 | 0.82664 | 0.85359 | 0.82652 | 0.85947 |

Note: the superscript * indicates that the model adopts dynamic feedback training strategy.